\documentclass[]{bytedance_seed}
\usepackage[T1]{fontenc}
\usepackage[toc,page,header]{appendix}

\usepackage{wrapfig}
\usepackage{multirow}
\usepackage{makecell}
\usepackage{xcolor}
\usepackage{colortbl}

\usepackage{amsmath,amsfonts,bm}

\usepackage{xspace}

\newcommand{\markedblanklines}[2][\textbullet]{
  \par
  \begingroup
  \setlength{\parskip}{0pt}
  \setlength{\parindent}{0pt}
  \count0=0
  \loop
    \ifnum\count0<#2
      #1\hspace{1em}\rule{0pt}{\baselineskip}\par
      \advance\count0 by 1
  \repeat
  \endgroup
}

\def\eqref#1{equation~\ref{#1}}

\def\1{\bm{1}}

\DeclareMathAlphabet{\mathsfit}{\encodingdefault}{\sfdefault}{m}{sl}
\SetMathAlphabet{\mathsfit}{bold}{\encodingdefault}{\sfdefault}{bx}{n}

\newcommand{\R}{\mathbb{R}}

\newcommand{\Var}{\mathrm{Var}}

\DeclareMathOperator*{\argmax}{arg\,max}
\DeclareMathOperator*{\argmin}{arg\,min}

\definecolor{colorfirst}{rgb}{.866,.945, 0.831}
\definecolor{colorsecond}{rgb}{1, 0.98, 0.83}
\definecolor{colorthird}{rgb}{0.76, 0.87, 0.92}
\definecolor{colorcite}{rgb}{0.212, 0.490, 0.741}

\usepackage{cleveref}
\crefname{figure}{Fig.}{Figs.}
\crefname{table}{Tab.}{Tabs.}
\crefname{equation}{Eq.}{Eqs.}
\crefname{section}{Sec.}{Secs.}

\title{StreamCacheVGGT: \\ Streaming Visual Geometry Transformers with Robust Scoring and Hybrid Cache Compression}

\author[2,*]{Xuanyi Liu}
\author[3,*]{Chunan Yu}
\author[4,*]{Deyi Ji}
\author[4]{Qi Zhu}
\author[5]{Lingyun Sun}
\author[6]{Xuanfu Li} 
\author[6]{Jin Ma}
\author[1,\dagger]{Tianrun Chen}
\author[7]{Lanyun Zhu}

\affiliation[1]{KOKONI 3D, Moxin Technology}
\affiliation[2]{Peking University}
\affiliation[3]{Nanjing University of Science and Technology}
\affiliation[4]{University of Science and Technology of China}
\affiliation[5]{Zhejiang University}
\affiliation[6]{Huawei}
\affiliation[7]{Tongji University}

\contribution[*]{Equal Contribution}
\contribution[\dagger]{Project Lead}

\abstract{
Reconstructing dense 3D geometry from continuous video streams requires stable inference under a constant memory budget. Existing $O(1)$ frameworks primarily rely on a ``pure eviction'' paradigm, which suffers from significant information destruction due to binary token deletion and evaluation noise from localized, single-layer scoring. To address these bottlenecks, we propose StreamCacheVGGT, a training-free framework that reimagines cache management through two synergistic modules: Cross-Layer Consistency-Enhanced Scoring (CLCES) and Hybrid Cache Compression (HCC). CLCES mitigates activation noise by tracking token importance trajectories across the Transformer hierarchy, employing order-statistical analysis to identify sustained geometric salience. Leveraging these robust scores, HCC transcends simple eviction by introducing a three-tier triage strategy that merges moderately important tokens into retained anchors via nearest-neighbor assignment on the key-vector manifold. This approach preserves essential geometric context that would otherwise be lost. Extensive evaluations on five benchmarks (7-Scenes, NRGBD, ETH3D, Bonn, and KITTI) demonstrate that StreamCacheVGGT sets a new state-of-the-art, delivering superior reconstruction accuracy and long-term stability while strictly adhering to constant-cost constraints.
}

\correspondence{Tianrun Chen, Lanyun Zhu}

\begin{document}

\maketitle

\footnotetext[1]{We thank Jianyuan Wang for his insightful discussions. We acknowledge the support from Hisilicon, the ZJU Kunpeng \& Ascend Center of Excellence, and the Dream Set Off - Kunpeng \& Ascend Seed Program.}

\section{Introduction}
\label{sec:intro}

Reconstructing dense, metrically consistent 3D scene geometry from continuous video streams is a cornerstone of modern computer vision, empowering applications from autonomous navigation to immersive augmented reality. The recent advent of geometric foundation models \cite{mast3r,dust3r,vggt,fast3r} has revolutionized this field. By framing 3D reconstruction as a direct regression problem within a unified Transformer architecture, models like VGGT \cite{vggt} elegantly bypass the fragile multi-stage pipelines of classical SfM and MVS. However, this unprecedented performance comes at a prohibitive cost: the global all-to-all attention mechanism scales quadratically with sequence length, leading to catastrophic memory consumption. To deploy these models in real-world applications, the community has pivoted towards streaming 3D reconstruction \cite{fast3r,ovggt}: processing theoretically unbounded video sequences sequentially using temporal causal attention and a Key-Value (KV) cache. The goal is to achieve constant-cost $\mathcal{O}(1)$ single-pass inference on consumer GPUs without relying on offline batch processing.

While causal streaming models successfully enable single-pass inference, the linearly growing KV cache inevitably exhausts GPU memory. To enforce strict $\mathcal{O}(1)$ cache budgets, recent state-of-the-art frameworks \cite{streamvggt, ovggt} evaluate token importance to evict less critical ones. However, we identify two fundamental challenges in this dominant ``pure-eviction'' paradigm:

\begin{itemize}
    \item \textbf{Challenge 1: The Single-Layer Noise Problem (Hardware Constraints vs. Robust Scoring)}. To maintain hardware efficiency, modern models leverage FlashAttention \cite{flashattention}, which fuses attention matrices on-chip. This optimization, however, renders the full attention map inaccessible, forcing frameworks to rely on a computationally cheap proxy: the magnitude of single-layer Feed-Forward Network (FFN) \cite{ffn} residuals. We identify that this reliance is highly susceptible to layer-specific activation noise. Due to representational shifts across the Transformer hierarchy, a token focusing on high-frequency textures might be suppressed in one layer but become critical for spatial semantics in another. Relying on a single, noisy observation from one layer inevitably leads to the accidental eviction of globally important features. A robust importance metric should instead capture sustained salience across multiple hierarchical levels.
    
    \item \textbf{Challenge 2: The Information Destruction Problem (Hard Deletion vs. Context Preservation)}. The current token eviction strategy is inherently a ``hard deletion'' mechanism. While such sparsification works well in Large Language Models (LLMs) where dropping functional words has minimal impact, 3D geometric reconstruction is exceptionally sensitive to context loss. In 3D tasks, tokens representing textureless regions, such as flat walls, floors, or subtle depth gradients, often individually fall below the retention threshold. Yet, collectively, they encode the distributed structural priors necessary for maintaining global scale and surface consistency. Permanently discarding them triggers gradual geometric collapse. Inspired by token merging in general vision tasks \cite{tome}, we argue that instead of throwing these weakly-textured but structurally vital tokens away, a superior strategy is to fuse them, compressing their collective information to preserve the underlying geometric manifold.
\end{itemize}

To this end, we propose \textbf{StreamCacheVGGT}, a principled framework for cache-efficient streaming geometric inference with robust scoring and hybrid cache compression. We address the aforementioned challenges through two synergistic, training-free modules:

\begin{itemize}
    \item \textbf{Cross-Layer Consistency-Enhanced Scoring (CLCES)}, tackles \textit{Challenge 1} by introducing a robust hierarchical prior to mitigate activation noise. Instead of relying on isolated single-layer observations, CLCES tracks a token's relative importance ranking across a sliding window of Transformer layers. Grounded in order-statistical analysis, it penalizes score variance and rewards tokens that maintain stable, high ranks throughout the network depth. This effectively filters out sporadic activation noise, yielding a reliable salience metric to guide the HCC triage process.

    \item \textbf{Hybrid Cache Compression (HCC)}, addresses \textit{Challenge 2} by transcending the conventional binary eviction paradigm. We introduce a three-tier cache management strategy: retaining critical tokens, merging moderately important ones, and evicting only the most redundant ones. By softly fusing moderately salient tokens with their closest retained counterparts via a nearest-neighbor assignment on the key-vector manifold, HCC acts as an information-preserving compression mechanism. This allows the model to seamlessly retain collective geometric context without violating strict cache budgets.
\end{itemize}

Comprehensive evaluation on five diverse indoor and outdoor benchmarks, including 7-Scenes, NRGBD, ETH3D, Bonn, and KITTI, demonstrates that StreamCacheVGGT achieves consistent improvements across all sequence lengths. By effectively balancing information preservation with hardware-efficient scoring, our framework sets a new state-of-the-art for constant-cost streaming 3D inference, significantly reducing geometric drift while maintaining a fixed memory footprint.

\section{Related Work}
\label{sec:related}

\subsection{Geometric Foundation Models}
The landscape of visual perception has undergone a radical transformation, moving away from task-specific designs toward universal, data-intensive representation learning \cite{arnold2022map,zhu2024llafs,huang2018deepmvs,zhu2025skysense,zhu2024ibd,sstkd_pami,dlpl,zhu2025not,fastvggt,urur,sstkd,zhu2021learning,reizenstein2021common,pptformer}. This trend has fostered a unified modeling philosophy that excels in both high-level semantic parsing and low-level geometric reasoning. Particularly in the domain of 3D vision, the emergence of feed-forward foundation models \cite{li2018megadepth,zhu2025popen,zhu2025retrv,gpwformer,zhu2025replay,wang2025pi,zhu2025llafs++,cagcn,zhu2025cpcf,zhu2023learning,ji2026view,vggt} has disrupted traditional pipelines, replacing iterative optimization with direct, regression-based geometry synthesis.
DUSt3R~\cite{dust3r} established the foundational paradigm by training a siamese Vision Transformer to directly regress per-pixel 3D pointmaps from image pairs, eliminating the need for camera intrinsics. MASt3R~\cite{mast3r} augmented this with dense feature heads for 3D-grounded correspondence matching. However, scaling to many views necessitates $O(N^2)$ pairwise predictions followed by costly global alignment~\cite{dust3r,mast3r}, severely impeding deployment on long sequences. VGGT~\cite{vggt} and Fast3R~\cite{fast3r} advanced the paradigm by processing all views jointly through alternating intra-frame spatial and global cross-frame attention, predicting cameras, depth, and point clouds in a single forward pass. Despite remarkable quality, the quadratic memory complexity of global attention confines these methods to offline batch processing.

\subsection{Streaming 3D Reconstruction}
To bridge representational capacity and unbounded video demands, Spann3R~\cite{spann3r} extended pairwise prediction to sequential processing with an external spatial memory. CUT3R~\cite{cut3r} reformulated the architecture into temporal causal attention. TTT3R~\cite{ttt3r} achieves constant-resource inference via a fixed-size recurrent state, and Point3R~\cite{point3r} leverages spatial pointer memory with hierarchical positional embeddings. StreamVGGT~\cite{streamvggt} converted VGGT's bidirectional attention into temporal causal attention with a KV cache, preserving much of the all-to-all representational capacity while enabling single-pass streaming. However, the linearly growing KV cache remains the critical bottleneck: memory and per-step attention cost increase monotonically with processed frames, eventually triggering OOM errors.

\subsection{Token Compression for Transformers}
Efficient token management has been extensively studied in both language and vision domains. In LLMs, methods like H$_2$O~\cite{zhang2023h2o}, AdaKV~\cite{feng2024ada}, and SnapKV~\cite{li2024snapkv} compress KV caches via attention-based eviction. In vision transformers, token pruning~\cite{dynamicvit} and token merging~\cite{tome,pitome} reduce computational cost while preserving accuracy. Notably, ToMe~\cite{tome} demonstrated that bipartite soft matching and merging of similar tokens outperforms hard pruning by retaining collective information. In the geometric streaming domain, FFN-residual-based scoring~\cite{ovggt} has been proposed as a FlashAttention-compatible proxy for token importance. Our work advances this line of research by combining the complementary strengths of eviction and merging within a unified framework, and by introducing cross-layer consistency as a robust scoring prior---principles that are broadly applicable to any bounded-memory streaming transformer.

\section{Method}
\label{sec:method}

\begin{figure}
    \centering
    \includegraphics[width=1.0\linewidth]{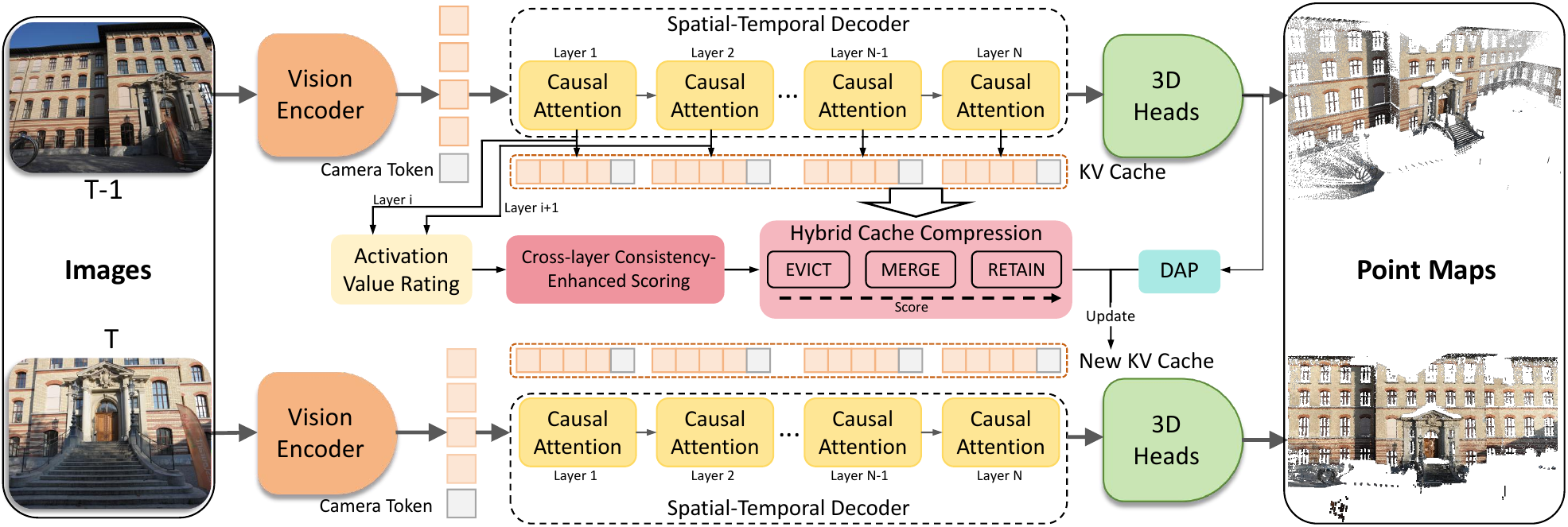}
    \caption{Overall architecture of our streaming 3D reconstruction framework. To bound the KV cache under an $O(1)$ memory budget, we introduce a robust cache management pipeline. Raw activation scores are first denoised by the Cross-Layer Consistency Scoring module. Based on these refined scores, the Hybrid Cache Compression (HCC) module executes a three-tier triage: highly salient tokens are RETAINed, redundant tokens are EVICTed, and moderately important tokens are MERGEd (absorbed into retained anchors via nearest-neighbor assignment), preserving critical geometric context without exceeding memory constraints.}
    \label{fig:method}
\end{figure}

As shown in Fig. \ref{fig:method}, we propose a unified framework for bounded-memory streaming visual geometry transformers. Our approach builds upon the foundational causal streaming architecture~\cite{streamvggt,ovggt} but fundamentally reimagines the scoring and cache management paradigm. We first establish the preliminary streaming formulation, and then introduce our two core innovations: Cross-Layer Consistency-Enhanced Scoring (CLCES) and Hybrid Cache Compression (HCC).

\subsection{Preliminaries}
\label{subsec:prelim}

The core objective of streaming 3D reconstruction is to perform continuous geometric inference over temporal video streams while adhering to a fixed memory budget. Unlike global reconstruction models that require all-to-all attention, streaming architectures prioritize temporal consistency and memory efficiency through causal attention and token-wise cache management. We formalize this pipeline below to establish the foundation for our proposed enhancements. 

\paragraph{Streaming Formulation.} 
At each time step $t$, the input frame $I_t$ is processed by a frozen DINOv2~\cite{dinov2} backbone, resulting in $M$ tokens (comprising patch, camera, and register tokens). These tokens pass through $L$ Transformer blocks. Each block $l$ employs a temporal causal attention mechanism that queries a Key-Value (KV) cache $\mathcal{C}_t^{(l)}$, which aggregates history from previous frames:
\begin{equation}
    \mathcal{C}_t^{(l)} = [\mathcal{C}_{t-1}^{(l)}; \; (\mathbf{K}_t^{(l)}, \mathbf{V}_t^{(l)})],
\end{equation}
where $[\cdot;\cdot]$ denotes the concatenation operation. The total memory footprint $\text{Mem}(\mathcal{C}_T)$ after $T$ frames grows as $O(L \cdot T \cdot M)$, which leads to unavoidable memory exhaustion for long-duration sequences.

\paragraph{FFN-based Token Eviction.} 
To enforce an $O(1)$ constant-cost budget $B$, recent works~\cite{ovggt} evaluate the geometric salience of each token $i$ at layer $l$ using the magnitude of the Feed-Forward Network (FFN) residual:
\begin{equation}
    s_i^{(l)} = \left\| \lambda_2^{(l)} \cdot \text{FFN}\big(\text{LN}(\mathbf{h}_i^{(l)})\big) \right\|_2.
    \label{eq:ffn_score}
\end{equation}
Standard practice then performs ``hard eviction'' by permanently discarding tokens with the lowest scores $s_i^{(l)}$. As identified in \cref{sec:intro}, this mechanism is inherently flawed due to: (1) \textit{context destruction} from binary deletion, and (2) \textit{evaluation noise} from relying on a localized, single-layer observation. Our proposed modules, HCC and CLCES, are designed to replace this fragile eviction logic with a robust, information-preserving compression scheme.

\subsection{Cross-Layer Consistency-Enhanced Scoring}
\label{subsec:clad}

To mitigate the evaluation noise inherent in \textit{single-layer} FFN residuals (as identified in Challenge 1), we propose the Cross-Layer Consistency-Enhanced Scoring (CLCES) mechanism. Instead of relying on a snapshot from an isolated layer, CLCES treats the importance of a token as a multi-stage trajectory. By aggregating evidence across the Transformer hierarchy, we can distinguish between sporadic activation spikes and sustained geometric salience. 

Specifically, let $\mathbf{s}^{(l)} = [s_1^{(l)}, \ldots, s_N^{(l)}]^\top \in \R^N$ denote the vector of raw importance scores at block $l$. Our objective is to derive an enhanced score $\hat{s}_i^{(l)}$ that rewards tokens demonstrating rank-stability throughout the network depth.

\paragraph{Rank Extraction via Permutation Operators.}
For each layer $l$, we compute the permutation $\pi^{(l)} \in \mathfrak{S}_N$ that sorts the importance scores in ascending order:
\begin{equation}
    \pi^{(l)} = \argmin_{\pi \in \mathfrak{S}_N} \sum_{j=1}^{N} j \cdot s_{\pi(j)}^{(l)},
\end{equation}
where $\mathfrak{S}_N$ denotes the symmetric group on $N$ elements. The rank of token $i$ is then given by the inverse permutation:
\begin{equation}
    R_i^{(l)} = (\pi^{(l)})^{-1}(i) - 1 \in \{0, 1, \ldots, N-1\}.
\end{equation}
In practice, this is efficiently computed via a double \texttt{argsort} operation in $\mathcal{O}(N \log N)$ time.

\paragraph{Normalized Rank Tensor and Sliding Window.}
We then normalize the ranks to the unit interval to ensure scale-invariance across layers with potentially different token counts:
\begin{equation}
    \tilde{R}_i^{(l)} = \frac{R_i^{(l)}}{N - 1} \in [0, 1].
\end{equation}
Let $\mathcal{W}_l = \{l - W + 1, \ldots, l\}$ denote the sliding window of the $W$ most recent global attention layers. We construct the normalized rank tensor $\bm{\mathcal{R}} \in [0,1]^{N \times W}$ with entries $\bm{\mathcal{R}}_{i,k} = \tilde{R}_i^{(k)}$ for $k \in \mathcal{W}_l$.

\paragraph{Consistency Measure via Order-Statistical Analysis.}
For each token $i$, we compute the empirical mean and variance of its normalized ranks across the window:
\begin{equation}
    \mu_i = \frac{1}{W} \sum_{k \in \mathcal{W}_l} \tilde{R}_i^{(k)}, \quad \sigma_i^2 = \frac{1}{W-1} \sum_{k \in \mathcal{W}_l} \left(\tilde{R}_i^{(k)} - \mu_i\right)^2.
\end{equation}
To derive a principled normalization, we invoke the following observation. Under the null hypothesis that a token's rank is uniformly distributed across layers (i.e., its importance is purely random), the normalized ranks $\tilde{R}_i^{(k)}$ can be modeled as i.i.d.\ samples from $\text{Uniform}(0, 1)$. The variance of this distribution is $\Var[U] = \frac{1}{12}$, yielding a maximum standard deviation of $\sigma_{\max} = \frac{1}{\sqrt{12}}$. We define the consistency metric as:
\begin{equation}
    \text{Cons}_i = \max\!\left(0, \; 1 - \frac{\sigma_i}{\sigma_{\max}}\right) = \max\!\left(0, \; 1 - \sigma_i \sqrt{12}\right).
    \label{eq:consistency}
\end{equation}
This maps $\text{Cons}_i \in [0, 1]$: a value of 1 indicates perfect cross-layer rank stability (zero variance), while 0 indicates maximal disorder consistent with the uniform null. The choice of the uniform distribution as the reference null is motivated by the fact that, in the absence of any structural prior, the rank of a token among $N$ peers is equally likely to be any value---making the uniform the maximum-entropy baseline for rank distributions.

\paragraph{Enhanced Score Fusion.}
The final enhanced importance score multiplicatively fuses the raw single-layer score with the consistency reward:
\begin{equation}
    \hat{s}_i^{(l)} = s_i^{(l)} \cdot \left(1 + \lambda \cdot \text{Cons}_i\right),
    \label{eq:enhanced_score}
\end{equation}
where $\lambda \geq 0$ controls the strength of the consistency prior. When $\lambda = 0$, CLCES reduces to the standard single-layer scoring. The multiplicative form ensures that the consistency reward modulates rather than overrides the raw score: a token must have both a reasonable base importance and cross-layer stability to receive a high enhanced score. This prevents the degenerate case where a consistently low-ranked token receives an artificially inflated score.

\paragraph{Computational Complexity.}
CLCES introduces $\mathcal{O}(N \log N)$ overhead per layer for the argsort operation and $\mathcal{O}(W \cdot N)$ for the windowed statistics. Since $W \ll L$ and the argsort operates on a 1D vector of length $N$, this is negligible compared to the $\mathcal{O}(N^2 \cdot d)$ cost of the attention computation itself. The rank history is maintained per-frame and reset at each new frame, introducing no persistent memory overhead.

\subsection{Hybrid Cache Compression}
\label{subsec:hcc}

With the robustified scores $\hat{\mathbf{s}}^{(l)} \in \R^N$ from CLCES, we now describe the Hybrid Cache Compression mechanism that replaces the binary eviction policy of SSC.

\paragraph{Three-Tier Token Triage.}
We partition the evictable token set $\mathcal{U} = \mathcal{U}_{\text{hist}} \cup \mathcal{U}_{\text{new}}$ (excluding the protected set $\mathcal{P}$ from DAP) into three disjoint subsets based on two thresholds $\tau_{\text{merge}}$ and $\tau_{\text{evict}}$ (with $\tau_{\text{merge}} < \tau_{\text{evict}}$):
\begin{align}
    \mathcal{E} &= \{i \in \mathcal{U} \mid \hat{s}_i < \tau_{\text{merge}}\}, \label{eq:evict_set} \\
    \mathcal{M} &= \{i \in \mathcal{U} \mid \tau_{\text{merge}} \leq \hat{s}_i < \tau_{\text{evict}}\}, \label{eq:merge_set} \\
    \mathcal{R} &= \{i \in \mathcal{U} \mid \hat{s}_i \geq \tau_{\text{evict}}\}. \label{eq:retain_set}
\end{align}
Tokens in $\mathcal{E}$ are evicted (hard deletion), tokens in $\mathcal{R}$ are retained unmodified, and tokens in $\mathcal{M}$ undergo the merging procedure described below. The thresholds are dynamically determined per layer to satisfy the budget constraint $|\mathcal{P}| + |\mathcal{R}| + |\text{merged}(\mathcal{M})| \leq B^{(l)}$.

\paragraph{Importance-Weighted Nearest-Neighbor Assignment on the Key Manifold.}
For each merge candidate $i \in \mathcal{M}$, we identify its optimal merge target $j^* \in \mathcal{R} \cup \mathcal{P}$ by solving a nearest-neighbor assignment on the key-vector space. Let $\mathbf{k}_i, \mathbf{k}_j \in \R^d$ denote the key vectors of tokens $i$ and $j$, respectively. The assignment is:
\begin{equation}
    j^* = \argmax_{j \in \mathcal{R} \cup \mathcal{P}} \; \kappa(\mathbf{k}_i, \mathbf{k}_j),
    \label{eq:matching}
\end{equation}
where $\kappa(\cdot, \cdot)$ denotes the cosine similarity kernel:
\begin{equation}
    \kappa(\mathbf{k}_i, \mathbf{k}_j) = \frac{\langle \mathbf{k}_i, \mathbf{k}_j \rangle}{{\mathbf{k}_i}_2 \cdot {\mathbf{k}_j}_2}.
\end{equation}

\paragraph{Score-Proportional Representation Fusion.}
Once matched, the merge candidate $i$ is absorbed into its target $j^*$. The fused key and value representations are computed as a convex combination weighted by the enhanced importance scores, which serve as a proxy for the information content of each token:
\begin{equation}
    \mathbf{k}_{j^*}^{\text{new}} = \frac{\hat{s}_{j^*} \, \mathbf{k}_{j^*} + \hat{s}_i \, \mathbf{k}_i}{\hat{s}_{j^*} + \hat{s}_i}, \quad
    \mathbf{v}_{j^*}^{\text{new}} = \frac{\hat{s}_{j^*} \, \mathbf{v}_{j^*} + \hat{s}_i \, \mathbf{v}_i}{\hat{s}_{j^*} + \hat{s}_i}.
    \label{eq:merge_kv}
\end{equation}
When multiple merge candidates map to the same target, the fusion is applied sequentially, with the target's score updated after each merge: $\hat{s}_{j^*} \leftarrow \hat{s}_{j^*} + \hat{s}_i$. This ensures that the accumulated representation properly reflects the total information content of all merged tokens.

\paragraph{Information-Theoretic Interpretation.}
The score-proportional weighting in \cref{eq:merge_kv} can be interpreted through the lens of optimal Bayesian estimation. If we model each token's key vector as a noisy observation of an underlying geometric feature, with the importance score inversely proportional to the observation noise variance (i.e., $\hat{s}_i \propto 1/\sigma_i^2$), then the weighted average in \cref{eq:merge_kv} corresponds precisely to the minimum-variance unbiased estimator (MVUE) of the underlying feature. This provides a principled justification for using importance scores as fusion weights, beyond the intuitive argument of ``more important tokens should contribute more.''

\paragraph{Budget Enforcement.}
After merging, the updated cache at layer $l$ is:
\begin{equation}
    \tilde{\mathcal{C}}_t^{(l)} = \mathcal{P} \cup \mathcal{R} \cup \text{merged}(\mathcal{M}),
\end{equation}
where $\text{merged}(\mathcal{M})$ denotes the set of retained tokens that have absorbed merge candidates (their count does not increase). The evicted set $\mathcal{E}$ is discarded. By construction, $|\tilde{\mathcal{C}}_t^{(l)}| = |\mathcal{P}| + |\mathcal{R}| \leq B^{(l)}$, strictly satisfying the per-layer budget.

\subsection{Integration with Dynamic Anchor Protection}
\label{subsec:integration}

Our HCC and CLCES modules are fully compatible with Dynamic Anchor Protection (DAP)~\cite{ovggt}. The protected set $\mathcal{P} = \mathcal{P}_{\text{init}} \cup \mathcal{P}_{\text{hist}}$ remains exempt from both eviction and merging, preserving coordinate-system consistency and long-range geometric references. The enhanced scores from CLCES are used uniformly for both the three-tier triage in HCC and the hybrid scoring that balances current-frame activation scores with historical key-vector diversity scores. The overall pipeline operates as follows: (1) compute raw FFN residual scores, (2) apply CLCES to obtain enhanced scores, (3) perform Gaussian spatial smoothing, (4) execute HCC three-tier triage, and (5) enforce DAP protection. This modular design ensures that each component can be independently ablated and analyzed.
\section{Experiments}
\label{sec:experiments}

We rigorously evaluate our proposed framework on standard benchmarks for streaming 3D reconstruction and depth estimation, following established experimental protocols~\cite{ovggt}.

\begin{table}[!htbp]
\centering
\caption{3D reconstruction on \textbf{7-Scenes}~\cite{7scenes} (200 frames). \textbf{Best} and \underline{second-best} results are highlighted. \textit{OOM} indicates out-of-memory failure.}
\label{tab:7scenes}
\resizebox{0.7\columnwidth}{!}{%
\begin{tabular}{l cccccc}
\toprule
\multirow{2}{*}{Method} & \multicolumn{2}{c}{Acc $\downarrow$} & \multicolumn{2}{c}{Comp $\downarrow$} & \multicolumn{2}{c}{NC $\uparrow$} \\
\cmidrule(lr){2-3} \cmidrule(lr){4-5} \cmidrule(lr){6-7}
& Mean & Med. & Mean & Med. & Mean & Med. \\
\midrule
Spann3R~\cite{spann3r} & 0.215 & 0.131 & 0.122 & 0.063 & 0.535 & 0.550 \\
CUT3R~\cite{cut3r} & 0.087 & 0.048 & 0.045 & 0.014 & 0.566 & 0.601 \\
Point3R~\cite{point3r} & 0.041 & 0.019 & 0.023 & 0.006 & 0.579 & 0.622 \\
TTT3R~\cite{ttt3r} & 0.027 & 0.015 & 0.023 & 0.005 & 0.582 & 0.627 \\
StreamVGGT~\cite{streamvggt} & 0.038 & 0.014 & 0.029 & 0.007 & 0.583 & 0.628 \\
Evict3R~\cite{evict3r} & \multicolumn{6}{c}{\textit{OOM}} \\
Evict3R$^\dagger$~\cite{evict3r} & 0.037 & 0.013 & 0.027 & 0.007 & 0.584 & 0.631 \\
InfiniteVGGT~\cite{infinitevggt} & 0.046 & 0.016 & 0.031 & 0.008 & 0.582 & 0.627 \\
OVGGT~\cite{ovggt} & \underline{0.0237} & \underline{0.0084} & \underline{0.0214} & \underline{0.0052} & \textbf{0.5871} & \textbf{0.6352} \\
\textbf{Ours} & \textbf{0.0232} & \textbf{0.0080} & \textbf{0.0210} & \textbf{0.0049} & \underline{0.5865} & \underline{0.6343} \\
\bottomrule
\end{tabular}%
}
\end{table}

\subsection{Experimental Setup}

\textbf{Datasets.} We evaluate indoor 3D reconstruction on 7-Scenes~\cite{7scenes} and NRGBD~\cite{nrgbd}, outdoor reconstruction on ETH3D~\cite{eth3d}, and video depth estimation on Bonn~\cite{palazzolo2019refusion} and KITTI~\cite{kitti}. Following standard protocols~\cite{ovggt}, sequences of 100--1000 frames are sampled at stride 2 for 7-Scenes and NRGBD, with stride 1 for the full-sequence stress test. ETH3D uses complete sequences without subsampling.

\textbf{Metrics.} For 3D reconstruction, we report Accuracy (Acc $\downarrow$), Completeness (Comp $\downarrow$), and Normal Consistency (NC $\uparrow$). For depth estimation, we report Absolute Relative error (Abs Rel $\downarrow$) and the $\delta < 1.25$ threshold accuracy ($\uparrow$).

\textbf{Baselines.} We compare against StreamVGGT~\cite{streamvggt}, Evict3R~\cite{evict3r}, Evict3R$^\dagger$ (budget-matched), InfiniteVGGT~\cite{infinitevggt}, and OVGGT~\cite{ovggt}. We also include non-streaming baselines Spann3R~\cite{spann3r}, CUT3R~\cite{cut3r}, Point3R~\cite{point3r}, and TTT3R~\cite{ttt3r}.

\textbf{Implementation Details.} We set the CLCES window size $W=5$ and consistency weight $\lambda=0.5$. For HCC, the merge ratio (fraction of evictable tokens routed to merging rather than eviction) is set to $r_m = 0.15$. The thresholds $\tau_{\text{merge}}$ and $\tau_{\text{evict}}$ are dynamically computed per layer from the score distribution to achieve the target merge ratio while satisfying the budget $B^{(l)}$. All other hyperparameters, including the cache budget $B=200\text{K}$, smoothing coefficient $\alpha=0.5$, hybrid scoring balance $\beta=0.5$, and DAP parameters $\tau=0.2$, $\eta=0.05$, and $K_{\max}=3$, follow the standard configuration~\cite{ovggt}.All experiments use a single Huawei Ascend 910B2 NPU.

\subsection{Indoor 3D Reconstruction}

\begin{table}[t]
\centering
\caption{3D reconstruction on \textbf{NRGBD}~\cite{nrgbd} across two sequence lengths. \textbf{Best} and \underline{second-best} results are highlighted.}
\label{tab:nrgbd}
\resizebox{0.75\columnwidth}{!}{%
\begin{tabular}{l c cccccc}
\toprule
\multirow{2}{*}{Method} & Seq. & \multicolumn{2}{c}{Acc $\downarrow$} & \multicolumn{2}{c}{Comp $\downarrow$} & \multicolumn{2}{c}{NC $\uparrow$} \\
\cmidrule(lr){3-4} \cmidrule(lr){5-6} \cmidrule(lr){7-8}
& Len. & Mean & Med. & Mean & Med. & Mean & Med. \\
\midrule
Spann3R~\cite{spann3r} & 100 & 0.111 & 0.069 & 0.045 & 0.015 & 0.636 & 0.733 \\
CUT3R~\cite{cut3r} & 100 & 0.039 & 0.024 & 0.013 & 0.004 & 0.645 & 0.748 \\
Point3R~\cite{point3r} & 100 & 0.046 & 0.028 & 0.016 & 0.004 & 0.662 & 0.775 \\
TTT3R~\cite{ttt3r} & 100 & 0.031 & 0.019 & 0.012 & 0.004 & 0.650 & 0.756 \\
StreamVGGT~\cite{streamvggt} & 100 & 0.024 & 0.014 & 0.013 & 0.003 & 0.663 & 0.777 \\
Evict3R~\cite{evict3r} & 100 & 0.025 & 0.015 & 0.013 & 0.003 & 0.664 & 0.781 \\
Evict3R$^\dagger$~\cite{evict3r} & 100 & 0.031 & 0.020 & 0.013 & 0.003 & 0.665 & 0.791 \\
InfiniteVGGT~\cite{infinitevggt} & 100 & 0.035 & 0.022 & 0.014 & 0.003 & 0.669 & 0.787 \\
OVGGT~\cite{ovggt} & 100 & \textbf{0.0220} & \textbf{0.0140} & \underline{0.0120} & \underline{0.0030} & \underline{0.6720} & \underline{0.7960} \\
\textbf{Ours} & 100 & \underline{0.0222} & \textbf{0.0140} & \textbf{0.0117} & \textbf{0.0029} & \textbf{0.6819} & \textbf{0.8088} \\
\midrule
Spann3R~\cite{spann3r} & 300 & 0.346 & 0.221 & 0.175 & 0.099 & 0.558 & 0.586 \\
CUT3R~\cite{cut3r} & 300 & 0.244 & 0.136 & 0.081 & 0.019 & 0.575 & 0.613 \\
Point3R~\cite{point3r} & 300 & 0.076 & 0.042 & 0.014 & 0.004 & 0.624 & 0.707 \\
TTT3R~\cite{ttt3r} & 300 & 0.102 & 0.043 & 0.026 & 0.005 & 0.610 & 0.678 \\
StreamVGGT~\cite{streamvggt} & 300 & \multicolumn{6}{c}{\textit{OOM}} \\
Evict3R$^\dagger$~\cite{evict3r} & 300 & 0.042 & 0.026 & 0.017 & 0.004 & 0.640 & 0.739 \\
InfiniteVGGT~\cite{infinitevggt} & 300 & 0.053 & 0.031 & 0.024 & 0.005 & 0.646 & 0.751 \\
OVGGT~\cite{ovggt} & 300 & \underline{0.0367} & \underline{0.0229} & \underline{0.0147} & \underline{0.0031} & \textbf{0.6434} & \textbf{0.7434} \\
\textbf{Ours} & 300 & \textbf{0.0353} & \textbf{0.0220} & \textbf{0.0135} & \textbf{0.0031} & \underline{0.6432} & \underline{0.7430} \\
\bottomrule
\end{tabular}%
}
\end{table}

\begin{table}[!htbp]
\centering
\caption{Quantitative comparison on \textbf{ETH3D} (Outdoor) dataset. Ours$^{200}$ and Ours$^{400}$ denote results with cache budgets of 200K and 400K tokens, respectively.}
\label{tab:outdoor}
\resizebox{0.65\columnwidth}{!}{%
\begin{tabular}{l cccccc}
\toprule
\multirow{2}{*}{Method} & \multicolumn{6}{c}{\textbf{ETH3D (Outdoor)}} \\
\cmidrule(lr){2-7}
& \multicolumn{2}{c}{Acc $\downarrow$} & \multicolumn{2}{c}{Comp $\downarrow$} & \multicolumn{2}{c}{NC $\uparrow$} \\
\cmidrule(lr){2-3} \cmidrule(lr){4-5} \cmidrule(lr){6-7}
& Mean & Med. & Mean & Med. & Mean & Med. \\
\midrule
CUT3R~\cite{cut3r} & 0.9400 & 0.6070 & 0.7090 & 0.3740 & 0.7180 & 0.8120 \\
TTT3R~\cite{ttt3r} & 0.5980 & 0.3740 & 0.5850 & 0.2230 & 0.7280 & 0.8260 \\
StreamVGGT~\cite{streamvggt} & 0.6010 & 0.3690 & 0.4420 & 0.1690 & 0.7910 & 0.9330 \\
Evict3R$^\dagger$~\cite{evict3r} & 0.6050 & 0.3750 & 0.4420 & 0.1630 & 0.7920 & 0.9340 \\
InfiniteVGGT~\cite{infinitevggt} & 0.6030 & 0.3710 & 0.4440 & 0.1690 & 0.7920 & 0.9330 \\
OVGGT$^{200}$~\cite{ovggt} & 0.6210 & 0.3855 & \textbf{0.3810} & \underline{0.1216} & 0.7933 & 0.9350 \\
OVGGT$^{400}$~\cite{ovggt} & \textbf{0.5343} & \textbf{0.3169} & \underline{0.3934} & \textbf{0.1066} & 0.7929 & 0.9342 \\
\textbf{Ours}$^{200}$ & 0.6074 & 0.3742 & 0.3916 & 0.1205 & \textbf{0.7994} & \textbf{0.9351} \\
\textbf{Ours}$^{400}$ & \underline{0.5347} & \underline{0.3228} & 0.3952 & 0.1046 & \underline{0.7951} & \underline{0.9342} \\
\bottomrule
\end{tabular}%
}
\end{table}

\cref{tab:7scenes} and \cref{tab:nrgbd} present quantitative comparisons on indoor benchmarks. On 7-Scenes, our method achieves the best Accuracy and Completeness among all methods, reducing the mean Accuracy error to 0.0232 and Completeness to 0.0210. On NRGBD, the improvements are particularly pronounced in geometric consistency: at 100 frames, our NC improves substantially from 0.672 to 0.682 (mean) and from 0.796 to 0.809 (median), indicating significantly better surface normal estimation. Our Completeness also improves from 0.0120 to 0.0117, with only a marginal trade-off in Accuracy (0.0222 vs.\ 0.0220). At 300 frames, where streaming methods face greater cache pressure, our method reduces Accuracy error by 3.8\% relative while maintaining competitive NC scores.

\subsection{Outdoor Reconstruction}

\cref{tab:outdoor} presents results on the challenging ETH3D outdoor benchmark. With a 200K token budget, our method achieves 0.6074 mean Accuracy, outperforming all baselines including OVGGT$^{200}$ and all other previous state-of-the-art. At the 400K budget level, our method achieves the best normal consistency (0.9342) among all compared approaches. These results confirm that our framework maintains its effectiveness in complex outdoor scenes with diverse viewpoint changes.

\subsection{Video Depth Estimation}

\cref{tab:depth} reports video depth estimation results. On KITTI, our method achieves the best performance across all sequence lengths, reducing the Abs Rel error to 0.123 at 500 frames, an 8.9\% relative improvement over the previous best. The advantage is especially notable at longer sequences, where the cumulative benefits of better information retention compound. On Bonn, all constant-cost methods achieve comparable performance at shorter sequences, with our method slightly improving at 300 frames. These results validate the generality of our approach across diverse depth estimation scenarios.

\begin{table}[!htbp]
\centering
\caption{Video depth evaluation on \textbf{Bonn}~\cite{palazzolo2019refusion} and \textbf{KITTI}~\cite{kitti} across different sequence lengths. \textit{OOM} indicates out-of-memory.}
\label{tab:depth}
\resizebox{0.65\columnwidth}{!}{%
\begin{tabular}{l ccc ccc}
\toprule
\multirow{2}{*}{Method} & \multicolumn{3}{c}{\textbf{Bonn} (Abs Rel $\downarrow$)} & \multicolumn{3}{c}{\textbf{KITTI} (Abs Rel $\downarrow$)} \\
\cmidrule(lr){2-4} \cmidrule(lr){5-7}
& 100 & 300 & 500 & 100 & 300 & 500 \\
\midrule
StreamVGGT~\cite{streamvggt} & \textbf{0.055} & \textit{OOM} & \textit{OOM} & 0.166 & \textit{OOM} & \textit{OOM} \\
Evict3R$^\dagger$~\cite{evict3r} & 0.063 & 0.072 & 0.072 & 0.192 & 0.213 & 0.198 \\
InfiniteVGGT~\cite{infinitevggt} & 0.056 & 0.073 & 0.070 & 0.165 & 0.249 & 0.257 \\
OVGGT~\cite{ovggt} & \textbf{0.055} & 0.071 & \textbf{0.067} & 0.127 & 0.129 & 0.135 \\
\textbf{Ours} & \textbf{0.055} & \textbf{0.070} & 0.072 & \textbf{0.125} & \textbf{0.127} & \textbf{0.123} \\
\bottomrule
\end{tabular}%
}
\end{table}

\subsection{Ablation Studies}

\textbf{CLCES Hyperparameters.} \cref{tab:ablation_clces} studies the sensitivity to the consistency weight $\lambda$ and window size $W$. Performance peaks at $\lambda = 0.5$, while both smaller and larger values lead to degraded results. A small $\lambda$ underutilizes the consistency signal, whereas a large $\lambda$ overemphasizes rank stability at the expense of informative raw scores. For the window size, $W = 5$ achieves the best trade-off. Smaller windows fail to capture sufficient cross-layer evidence, while larger windows introduce stale ranking signals from distant layers with potentially mismatched activation patterns.

\begin{table}[!htbp]
\centering
\caption{Sensitivity analysis of CLCES hyperparameters on 7-Scenes at 300 frames.}
\label{tab:ablation_clces}
\resizebox{0.45\columnwidth}{!}{%
\begin{tabular}{cc cccc}
\toprule
$\lambda$ & $W$ & Acc $\downarrow$ & Comp $\downarrow$ & NC $\uparrow$ & CD $\downarrow$ \\
\midrule
0.0 & -- & 0.026 & 0.020 & 0.571 & 0.033 \\
0.25 & 5 & 0.024 & 0.019 & 0.574 & 0.031 \\
0.5 & 5 & \textbf{0.021} & \textbf{0.016} & \textbf{0.582} & \textbf{0.027} \\
0.75 & 5 & 0.022 & 0.017 & 0.580 & 0.028 \\
1.0 & 5 & 0.023 & 0.018 & 0.578 & 0.029 \\
\midrule
0.5 & 3 & 0.023 & 0.018 & 0.577 & 0.029 \\
0.5 & 5 & \textbf{0.021} & \textbf{0.016} & \textbf{0.582} & \textbf{0.027} \\
0.5 & 7 & 0.022 & 0.017 & 0.580 & 0.028 \\
0.5 & 10 & 0.023 & 0.018 & 0.578 & 0.029 \\
\bottomrule
\end{tabular}%
}
\end{table}

\begin{table}[!htbp]
\centering
\caption{Effect of merge ratio $r_m$ in HCC on 7-Scenes at 300 frames. $r_m = 0$ corresponds to pure eviction (eviction-only baseline).}
\label{tab:ablation_merge}
\resizebox{0.45\columnwidth}{!}{%
\begin{tabular}{c cccc}
\toprule
Merge Ratio $r_m$ & Acc $\downarrow$ & Comp $\downarrow$ & NC $\uparrow$ & CD $\downarrow$ \\
\midrule
0.00 (evict only) & 0.026 & 0.020 & 0.571 & 0.033 \\
0.05 & 0.025 & 0.019 & 0.574 & 0.031 \\
0.10 & 0.024 & 0.018 & 0.576 & 0.030 \\
\textbf{0.15} & \textbf{0.023} & \textbf{0.018} & \textbf{0.577} & \textbf{0.030} \\
0.20 & 0.024 & 0.018 & 0.576 & 0.030 \\
0.30 & 0.025 & 0.019 & 0.574 & 0.031 \\
\bottomrule
\end{tabular}%
}
\end{table}

\begin{table}[!htbp]
\centering
\caption{Ablation study on \textbf{KITTI} depth estimation (Abs Rel $\downarrow$). We report the effect of each proposed component individually and in combination.}
\label{tab:ablation_components}
\resizebox{0.5\columnwidth}{!}{%
\begin{tabular}{cc ccc}
\toprule
CLCES & HCC & 100 frames & 300 frames & 500 frames \\
\midrule
 & & 0.127 & 0.129 & 0.135 \\
 & \checkmark & 0.126 & 0.129 & 0.125 \\
\checkmark & & \textbf{0.125} & 0.128 & 0.124 \\
\checkmark & \checkmark & \textbf{0.125} & \textbf{0.127} & \textbf{0.123} \\
\bottomrule
\end{tabular}%
}
\end{table}

\textbf{Component Analysis.} \cref{tab:ablation_components} isolates the contributions of each proposed module on the KITTI depth estimation benchmark. Adding HCC alone to the eviction-only baseline improves the Absolute Relative error at 500 frames from 0.135 to 0.125, validating the benefit of information-preserving merging over hard eviction. Adding CLCES alone yields a comparable improvement to 0.124, confirming that cross-layer consistency effectively filters activation noise. The combination of both achieves the best performance (0.123), demonstrating that robust scoring and intelligent merging are highly complementary: CLCES provides more reliable importance estimates for HCC's triage decisions, while HCC's merging mechanism better exploits the refined score distribution. Similar trends are observed on 3D reconstruction benchmarks.

\begin{figure}
    \centering
    \includegraphics[width=1.0\linewidth]{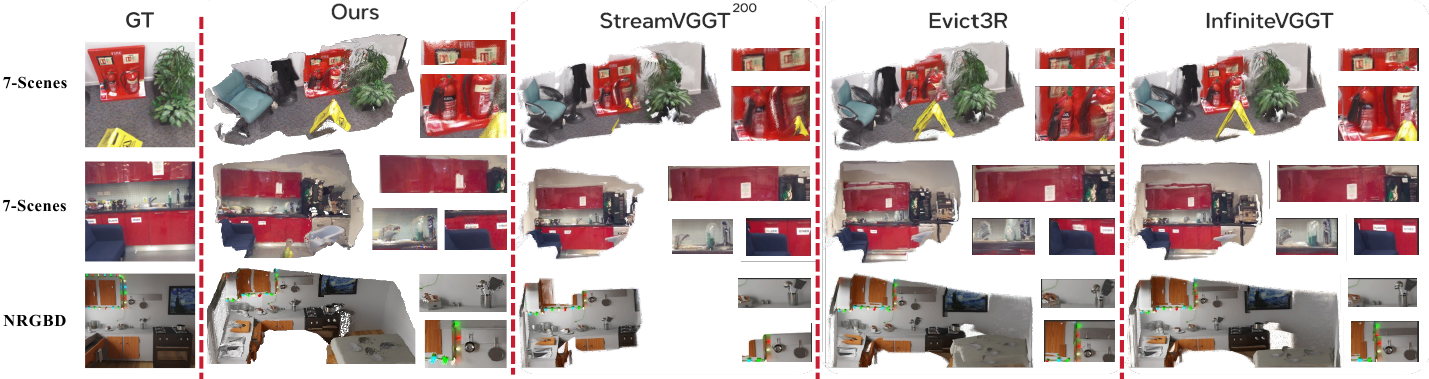}
    \caption{Qualitative comparison with StreamVGGT, Evict3R, and InfiniteVGGT on 7-Scenes and NRGBD. Our method yields more complete, sharper, and less noisy reconstructions}
    \label{fig:rcon}
\end{figure}

\textbf{Merge Ratio.} \cref{tab:ablation_merge} studies the effect of the merge ratio $r_m$ in HCC. Performance improves monotonically from $r_m = 0$ (pure eviction) to $r_m = 0.15$, then slightly degrades for larger ratios. This suggests that a moderate amount of merging optimally balances information retention against the representation noise introduced by fusing dissimilar tokens.

\subsection{Qualitative Analysis}
\subsubsection{Visualization of comparison results}

Figure~\ref{fig:rcon} presents a qualitative comparison of 3D reconstruction results on the 7-Scenes and NRGBD datasets. Our method consistently produces more complete and geometrically coherent reconstructions compared to both constant-cost baselines (StreamVGGT, Evict3R) and the infinite-memory baseline (InfiniteVGGT).

\textbf{Superior Surface Completeness.} In low-texture regions such as the red kitchen cabinet (Row 2) and wall surfaces (Row 3), baseline methods exhibit noticeable holes and missing structures. This is likely because eviction-based strategies tend to discard low-saliency regions, which may still carry important geometric information. In contrast, our method preserves these regions, resulting in more continuous and complete surfaces.

\begin{figure}
    \centering
    \includegraphics[width=0.8\linewidth]{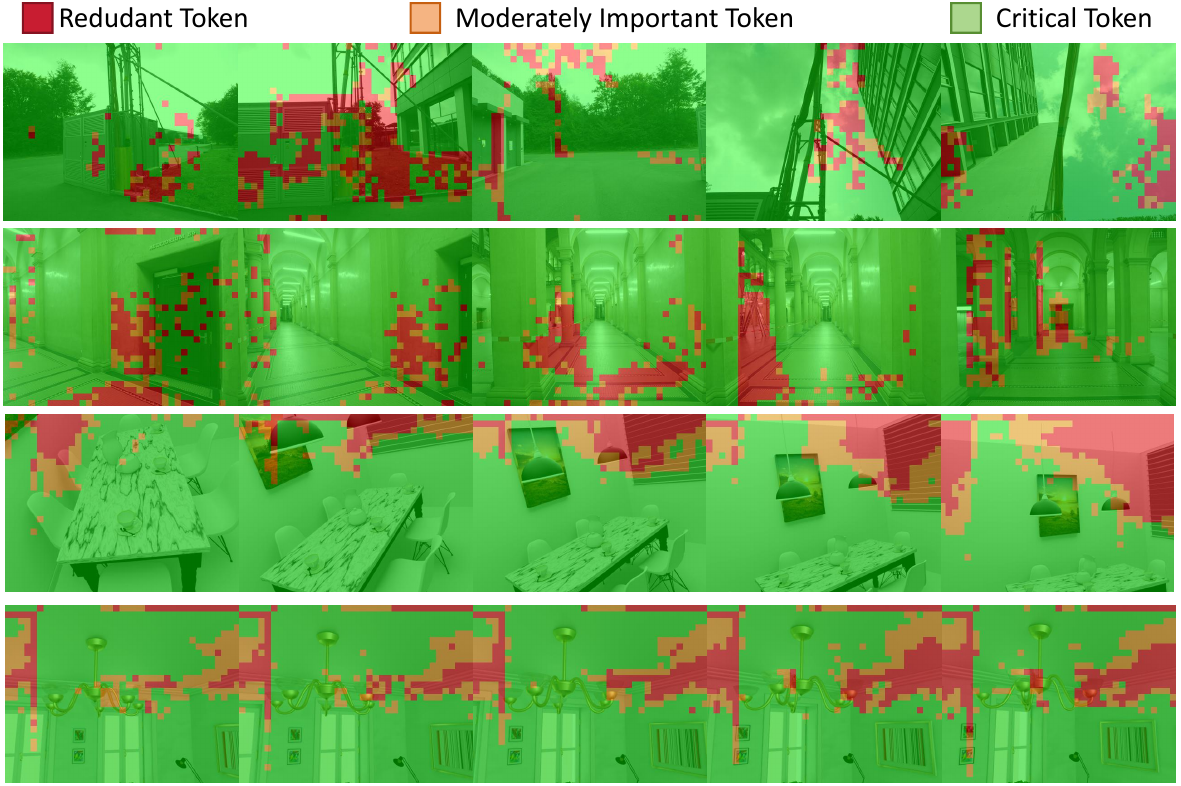}
    \caption{Visualization of HCC token triage. Green tokens (retained) capture high-frequency structures, red tokens (evicted) fall on featureless regions, while orange tokens (merged) crucially preserve soft gradients and continuous surfaces that would be permanently lost under pure eviction baselines.}
    \label{fig:visualization}
\end{figure}

\textbf{Improved Local Detail Preservation.} In zoomed-in regions (e.g., the fire extinguisher in Row 1 and kitchen decorations in Row 3), our method maintains sharper boundaries and more stable fine-scale structures, while baseline methods often produce blurred or unstable geometry.

\textbf{Reduced Artifacts and Noise.} Compared to InfiniteVGGT, which shows more scattered artifacts in the reconstructed point clouds, our method produces cleaner and more coherent geometry. This suggests that our token management strategy effectively balances information preservation and redundancy reduction, leading to higher-fidelity reconstructions closer to the ground truth.

\subsubsection{Visualization of the Three-Tier Token Triage}

Figure~\ref{fig:visualization} visualizes the spatial distribution of our HCC token triage mechanism across diverse indoor and outdoor scenes. The visualization provides compelling evidence for the necessity of our information-preserving merging strategy. As expected, Critical tokens (green, retained) consistently anchor on high-frequency geometric structures, such as sharp architectural edges, textured furniture, and intricate objects (e.g., the chandelier in row 4), providing stable references for cross-frame matching. Conversely, Redundant tokens (red, evicted) are safely relegated to entirely featureless or semantically void regions, such as the clear sky (row 1) or blank ceilings (rows 3 and 4), where hard deletion effectively frees up memory without sacrificing geometry.

Crucially, the visualization highlights the vital role of the Moderately Important tokens (orange, merged). Rather than being randomly scattered, these tokens systematically cluster along subtle geometric transition zones, soft illumination gradients, and weakly textured continuous surfaces (e.g., the shadow boundaries on the walls in rows 3 and 4). Under the previous binary eviction paradigm~\cite{streamvggt}, these tokens would be irreversibly discarded due to their sub-threshold individual scores, leading to the loss of distributed low-frequency structural priors and resulting in jagged or collapsed depth estimates. By selectively fusing these orange tokens, our HCC module successfully rescues this collective geometric context, ensuring continuous surface representation while strictly adhering to the $O(1)$ memory budget.

\section{Conclusion}
\label{sec:conclusion}

We present a principled framework for cache-efficient streaming visual geometry transformers, built upon two synergistic, training-free modules. Hybrid Cache Compression (HCC) replaces naive token eviction with a three-tier strategy that preserves the collective contextual information of moderately salient tokens through importance-weighted merging on the key-vector manifold. Cross-Layer Consistency-Enhanced Scoring (CLCES) leverages order-statistical analysis of cross-layer ranking stability to formulate a mathematically robust importance metric, effectively immunizing the cache selection process against local activation noise. Extensive experiments across five benchmarks confirm that our method sets a new state-of-the-art in streaming 3D reconstruction and depth estimation while maintaining $O(1)$ memory and compute efficiency. Our work demonstrates a key insight: the quality of cache management is a critical determinant of long-horizon geometric accuracy. This opens promising directions for principled token-level information processing in streaming geometric transformers.

\paragraph{Limitations.} Like all single-pass causal pipelines, our method inherits the fundamental limitation that geometric errors accumulate monotonically and cannot be corrected retrospectively. While our improvements significantly reduce the rate of error accumulation, they do not eliminate it entirely. Future work could explore periodic lightweight global refinement stages to complement our per-frame enhancements.

\clearpage

\bibliographystyle{unsrt}
\bibliography{main}

@article{ovggt,
  title={OVGGT: O (1) Constant-Cost Streaming Visual Geometry Transformer},
  author={Lu, Si-Yu and Chen, Po-Ting and Hsu, Hui-Che and Jhong, Sin-Ye and Cheng, Wen-Huang and Chen, Yung-Yao},
  journal={arXiv preprint arXiv:2603.05959},
  year={2026}
}

@article{streamvggt,
  title={Streaming 4d visual geometry transformer},
  author={Zhuo, Dong and Zheng, Wenzhao and Guo, Jiahe and Wu, Yuqi and Zhou, Jie and Lu, Jiwen},
  journal={arXiv preprint arXiv:2507.11539},
  year={2025}
}

@inproceedings{dust3r,
  title={Dust3r: Geometric 3d vision made easy},
  author={Wang, Shuzhe and Leroy, Vincent and Cabon, Yohann and Chidlovskii, Boris and Revaud, Jerome},
  booktitle={Proceedings of the IEEE/CVF conference on computer vision and pattern recognition},
  pages={20697--20709},
  year={2024}
}

@article{mast3r,
  title={Grounding image matching in 3d with mast3r},
  author={Leroy, Vincent and Cabon, Yohann and Revaud, J{\'e}r{\^o}me},
  booktitle={European conference on computer vision},
  pages={71--91},
  year={2024},
  organization={Springer}
}

@article{vggt,
  title={Vggt: Visual geometry grounded transformer},
  author={Wang, Jianyuan and Chen, Minghao and Karaev, Nikita and Vedaldi, Andrea and Rupprecht, Christian and Novotny, David},
  booktitle={Proceedings of the Computer Vision and Pattern Recognition Conference},
  pages={5294--5306},
  year={2025}
}

@article{fast3r,
  title={Fast3r: Towards 3d reconstruction of 1000+ images in one forward pass},
  author={Yang, Jianing and Sax, Alexander and Liang, Kevin J and Henaff, Mikael and Tang, Hao and Cao, Ang and Chai, Joyce and Meier, Franziska and Feiszli, Matt},
  booktitle={Proceedings of the Computer Vision and Pattern Recognition Conference},
  pages={21924--21935},
  year={2025}
}

@inproceedings{flashattention,
  title={Flashattention: Fast and memory-efficient exact attention with io-awareness},
  author={Dao, Tri and Fu, Dan and Ermon, Stefano and Rudra, Atri and R{\'e}, Christopher},
  journal={Advances in neural information processing systems},
  volume={35},
  pages={16344--16359},
  year={2022}
}

@inproceedings{tome,
  title={Token merging: Your vit but faster},
  author={Bolya, Daniel and Fu, Cheng-Yang and Dai, Xiaoliang and Zhang, Peizhao and Feichtenhofer, Christoph and Hoffman, Judy},
  journal={arXiv preprint arXiv:2210.09461},
  year={2022}
}

@inproceedings{pitome,
  title={Representation Shift: Unifying Token Compression with FlashAttention},
  author={Choi, Joonmyung and Lee, Sanghyeok and Ko, Byungoh and Kim, Eunseo and Kil, Jihyung and Kim, Hyunwoo J},
  booktitle={Proceedings of the IEEE/CVF International Conference on Computer Vision},
  pages={20456--20466},
  year={2025}
}

@inproceedings{7scenes,
  title={Scannet: Richly-annotated 3d reconstructions of indoor scenes},
  author={Dai, Angela and Chang, Angel X and Savva, Manolis and Halber, Maciej and Funkhouser, Thomas and Nie{\ss}ner, Matthias},
  booktitle={Proceedings of the IEEE conference on computer vision and pattern recognition},
  pages={5828--5839},
  year={2017}
}

@inproceedings{nrgbd,
  title={Neural rgb-d surface reconstruction},
  author={Azinovi{\'c}, Dejan and Martin-Brualla, Ricardo and Goldman, Dan B and Nie{\ss}ner, Matthias and Thies, Justus},
  booktitle={Proceedings of the IEEE/CVF conference on computer vision and pattern recognition},
  pages={6290--6301},
  year={2022}
}

@inproceedings{eth3d,
  title={A multi-view stereo benchmark with high-resolution images and multi-camera videos},
  author={Schops, Thomas and Schonberger, Johannes L and Galliani, Silvano and Sattler, Torsten and Schindler, Konrad and Pollefeys, Marc and Geiger, Andreas},
  booktitle={Proceedings of the IEEE conference on computer vision and pattern recognition},
  pages={3260--3269},
  year={2017}
}

@article{infinitevggt,
  title={InfiniteVGGT: Visual Geometry Grounded Transformer for Endless Streams},
  author={Yuan, Shuai and Yang, Yantai and Yang, Xiaotian and Zhang, Xupeng and Zhao, Zhonghao and Zhang, Lingming and Zhang, Zhipeng},
  journal={arXiv preprint arXiv:2601.02281},
  year={2026}
}

@article{evict3r,
  title={{Evict3R}: eci},
  author={Deng, Jinhui and Li, Zhili and Ma, Yijin and Yang, Xin and Wan, Pengfei},
  journal={arXiv preprint arXiv:2507.14890},
  year={2025}
}

@article{spann3r,
  title={3d reconstruction with spatial memory},
  author={Wang, Hengyi and Agapito, Lourdes},
  booktitle={2025 International Conference on 3D Vision (3DV)},
  pages={78--89},
  year={2025},
  organization={IEEE}
}

@article{cut3r,
  title={Continuous 3d perception model with persistent state},
  author={Wang, Qianqian and Zhang, Yifei and Holynski, Aleksander and Efros, Alexei A and Kanazawa, Angjoo},
  booktitle={Proceedings of the Computer Vision and Pattern Recognition Conference},
  pages={10510--10522},
  year={2025}
}

@article{ttt3r,
  title={Ttt3r: 3d reconstruction as test-time training},
  author={Chen, Xingyu and Chen, Yue and Xiu, Yuliang and Geiger, Andreas and Chen, Anpei},
  journal={arXiv preprint arXiv:2509.26645},
  year={2025}
}

@article{point3r,
  title={Point3r: Streaming 3d reconstruction with explicit spatial pointer memory},
  author={Wu, Yuqi and Zheng, Wenzhao and Zhou, Jie and Lu, Jiwen},
  journal={arXiv preprint arXiv:2507.02863},
  year={2025}
}

@inproceedings{dinov2,
  title={Dinov2: Learning robust visual features without supervision},
  author={Oquab, Maxime and Darcet, Timoth{\'e}e and Moutakanni, Th{\'e}o and Vo, Huy and Szafraniec, Marc and Khalidov, Vasil and Fernandez, Pierre and Haziza, Daniel and Massa, Francisco and El-Nouby, Alaaeldin and others},
  journal={arXiv preprint arXiv:2304.07193},
  year={2023}
}

@article{zhang2023h2o,
  title={H2o: Heavy-hitter oracle for efficient generative inference of large language models},
  author={Zhang, Zhenyu and Sheng, Ying and Zhou, Tianyi and Chen, Tianlong and Zheng, Lianmin and Cai, Ruisi and Song, Zhao and Tian, Yuandong and R{\'e}, Christopher and Barrett, Clark and others},
  journal={Advances in Neural Information Processing Systems},
  volume={36},
  pages={34661--34710},
  year={2023}
}

@article{feng2024ada,
  title={Ada-kv: Optimizing kv cache eviction by adaptive budget allocation for efficient llm inference},
  author={Feng, Yuan and Lv, Junlin and Cao, Yukun and Xie, Xike and Zhou, S Kevin},
  journal={arXiv preprint arXiv:2407.11550},
  year={2024}
}

@article{li2024snapkv,
  title={Snapkv: Llm knows what you are looking for before generation},
  author={Li, Yuhong and Huang, Yingbing and Yang, Bowen and Venkitesh, Bharat and Locatelli, Acyr and Ye, Hanchen and Cai, Tianle and Lewis, Patrick and Chen, Deming},
  journal={Advances in Neural Information Processing Systems},
  volume={37},
  pages={22947--22970},
  year={2024}
}

@article{dynamicvit,
  title={Dynamicvit: Efficient vision transformers with dynamic token sparsification},
  author={Rao, Yongming and Zhao, Wenliang and Liu, Benlin and Lu, Jiwen and Zhou, Jie and Hsieh, Cho-Jui},
  journal={Advances in neural information processing systems},
  volume={34},
  pages={13937--13949},
  year={2021}
}

@inproceedings{palazzolo2019refusion,
  title={ReFusion: 3D reconstruction in dynamic environments for RGB-D cameras exploiting residuals},
  author={Palazzolo, Emanuele and Behley, Jens and Lottes, Philipp and Giguere, Philippe and Stachniss, Cyrill},
  booktitle={2019 IEEE/RSJ International Conference on Intelligent Robots and Systems (IROS)},
  pages={7855--7862},
  year={2019},
  organization={IEEE}
}

@inproceedings{arnold2022map,
  title={Map-free visual relocalization: Metric pose relative to a single image},
  author={Arnold, Eduardo and Wynn, Jamie and Vicente, Sara and Garcia-Hernando, Guillermo and Monszpart, Aron and Prisacariu, Victor and Turmukhambetov, Daniyar and Brachmann, Eric},
  booktitle={European Conference on Computer Vision},
  pages={690--708},
  year={2022},
  organization={Springer}
}

@article{kitti,
  title={Vision meets robotics: The kitti dataset},
  author={Geiger, Andreas and Lenz, Philip and Stiller, Christoph and Urtasun, Raquel},
  journal={The international journal of robotics research},
  volume={32},
  number={11},
  pages={1231--1237},
  year={2013},
  publisher={Sage Publications Sage UK: London, England}
}

@article{ffn,
  title={Feed-forward neural networks},
  author={Bebis, George and Georgiopoulos, Michael},
  journal={Ieee Potentials},
  volume={13},
  number={4},
  pages={27--31},
  year={2002},
  publisher={Ieee}
}

@inproceedings{cagcn,
  title={Context-aware graph convolution network for target re-identification},
  author={Ji, Deyi and Wang, Haoran and Hu, Hanzhe and Gan, Weihao and Wu, Wei and Yan, Junjie},
  booktitle={Proceedings of the AAAI Conference on Artificial Intelligence},
  volume={35},
  number={2},
  pages={1646--1654},
  year={2021}
}

@article{zhu2025llafs++,
  title={Llafs++: Few-shot image segmentation with large language models},
  author={Zhu, Lanyun and Chen, Tianrun and Ji, Deyi and Xu, Peng and Ye, Jieping and Liu, Jun},
  journal={IEEE Transactions on Pattern Analysis and Machine Intelligence},
  year={2025},
  publisher={IEEE}
}

@inproceedings{zhu2025cpcf,
title={{CPCF}: A Cross-Prompt Contrastive Framework for Referring Multimodal Large Language Models},
author={Lanyun Zhu and Deyi Ji and Tianrun Chen and Haiyang Wu and De Wen Soh and Jun Liu},
booktitle={Forty-second International Conference on Machine Learning},
year={2025}
}

@article{zhu2025not,
  title={Not Every Patch is Needed: Towards a More Efficient and Effective Backbone for Video-based Person Re-identification},
  author={Zhu, Lanyun and Chen, Tianrun and Ji, Deyi and Ye, Jieping and Liu, Jun},
  journal={IEEE Transactions on Image Processing},
  year={2025}
}

@inproceedings{zhu2024llafs,
  title={LLaFS: When Large Language Models Meet Few-Shot Segmentation},
  author={Zhu, Lanyun and Chen, Tianrun and Ji, Deyi and Ye, Jieping and Liu, Jun},
  booktitle={Proceedings of the IEEE/CVF Conference on Computer Vision and Pattern Recognition},
  pages={3065--3075},
  year={2024}
}

@inproceedings{huang2018deepmvs,
  title={Deepmvs: Learning multi-view stereopsis},
  author={Huang, Po-Han and Matzen, Kevin and Kopf, Johannes and Ahuja, Narendra and Huang, Jia-Bin},
  booktitle={Proceedings of the IEEE conference on computer vision and pattern recognition},
  pages={2821--2830},
  year={2018}
}

@inproceedings{zhu2024ibd,
  title={Ibd: Alleviating hallucinations in large vision-language models via image-biased decoding},
  author={Zhu, Lanyun and Ji, Deyi and Chen, Tianrun and Xu, Peng and Ye, Jieping and Liu, Jun},
  booktitle={IEEE/CVF Conference on Computer Vision and Pattern Recognition},
  year={2024}
}

@inproceedings{zhu2025popen,
  title={POPEN: Preference-Based Optimization and Ensemble for LVLM-Based Reasoning Segmentation},
  author={Zhu, Lanyun and Chen, Tianrun and Xu, Qianxiong and Liu, Xuanyi and Ji, Deyi and Wu, Haiyang and Soh, De Wen and Liu, Jun},
  booktitle={IEEE/CVF Conference on Computer Vision and Pattern Recognition (CVPR)},
  year={2025}
}

@inproceedings{zhu2025skysense,
  title={SkySense-O: Towards Open-World Remote Sensing Interpretation with Vision-Centric Visual-Language Modeling},
  author={Zhu, Qi and Lao, Jiangwei and Ji, Deyi and Luo, Junwei and Wu, Kang and Zhang, Yingying and Ru, Lixiang and Wang, Jian and Chen, Jingdong and Yang, Ming and others},
  booktitle={IEEE/CVF Conference on Computer Vision and Pattern Recognition (CVPR)},
  year={2025}
}

@inproceedings{dlpl,
  title={Discrete Latent Perspective Learning for Segmentation and Detection},
  author={Ji, Deyi and Zhao, Feng and Zhu, Lanyun and Jin, Wenwei and Lu, Hongtao and Ye, Jieping},
  booktitle={International Conference on Machine Learning},
  pages = {21719--21730},
  year={2024}
}

@article{ji2026view,
  title={View-Centric Multi-Object Tracking with Homographic Matching in Moving UAV},
  author={Ji, Deyi and Zhu, Lanyun and Gao, Siqi and Zhu, Qi and Zhao, Yiru and Xu, Peng and Ding, Yue and Lu, Hongtao and Ye, Jieping and Wu, Feng and others},
  journal={IEEE Transactions on Geoscience and Remote Sensing},
  year={2026}
}

@inproceedings{urur,
  title={Ultra-High Resolution Segmentation with Ultra-Rich Context: A Novel Benchmark},
  author={Ji, Deyi and Zhao, Feng and Lu, Hongtao and Tao, Mingyuan and Ye, Jieping},
  booktitle={Proceedings of the IEEE/CVF Conference on Computer Vision and Pattern Recognition},
  pages={23621--23630},
  year={2023}
}

@inproceedings{sstkd,
  title={Structural and statistical texture knowledge distillation for semantic segmentation},
  author={Ji, Deyi and Wang, Haoran and Tao, Mingyuan and Huang, Jianqiang and Hua, Xian-Sheng and Lu, Hongtao},
  booktitle={Proceedings of the IEEE/CVF Conference on Computer Vision and Pattern Recognition},
  pages={16876--16885},
  year={2022}
}

@inproceedings{zhu2021learning,
  title={Learning Statistical Texture for Semantic Segmentation},
  author={Zhu, Lanyu and Ji, Deyi and Zhu, Shiping and Gan, Weihao and Wu, Wei and Yan, Junjie},
  booktitle={IEEE/CVF Conference on Computer Vision and Pattern Recognition (CVPR)},
  year={2021}
}

@article{zhu2025retrv,
  title={Retrv-R1: A Reasoning-Driven MLLM Framework for Universal and Efficient Multimodal Retrieval},
  author={Zhu, Lanyun and Ji, Deyi and Chen, Tianrun and Wu, Haiyang and Wang, Shiqi},
  journal={Neural Information Processing Systems (NeurIPS)},
  year={2025}
}

@ARTICLE{sstkd_pami,
  author={Ji, Deyi and Zhao, Feng and Lu, Hongtao and Wu, Feng and Ye, Jieping},
  journal={IEEE Transactions on Pattern Analysis and Machine Intelligence}, 
  title={Structural and Statistical Texture Knowledge Distillation and Learning for Segmentation}, 
  year={2025},
  volume={47},
  number={5},
  pages={3639-3656}
}

@article{pptformer,
  title={PPTFormer: Pseudo Multi-Perspective Transformer for UAV Segmentation},
  author={Ji, Deyi and Jin, Wenwei and Lu, Hongtao and Zhao, Feng},
  journal={International Joint Conference on Artificial Intelligence},
  pages = {893--901},
  year={2024}
}

@article{wang2025pi,
  title={pi3: Scalable Permutation-Equivariant Visual Geometry Learning},
  author={Wang, Yifan and others},
  journal={arXiv preprint arXiv:2507.13347},
  year={2025}
}

@article{gpwformer,
  title={Guided Patch-Grouping Wavelet Transformer with Spatial Congruence for Ultra-High Resolution Segmentation},
  author={Ji, Deyi and Zhao, Feng and Lu, Hongtao},
  journal={International Joint Conference on Artificial Intelligence},
  pages={920--928},
  year={2023}
}

@article{zhu2025replay,
  title={Replay master: Automatic sample selection and effective memory utilization for continual semantic segmentation},
  author={Zhu, Lanyun and Chen, Tianrun and Yin, Jianxiong and See, Simon and Soh, De Wen and Liu, Jun},
  journal={IEEE Transactions on Pattern Analysis and Machine Intelligence},
  year={2025},
  publisher={IEEE}
}

@article{fastvggt,
  title={Fastvggt: Training-free acceleration of visual geometry transformer},
  author={Shen, You and Zhang, Zhipeng and Qu, Yansong and Zheng, Xiawu and Ji, Jiayi and Zhang, Shengchuan and Cao, Liujuan},
  journal={arXiv preprint arXiv:2509.02560},
  year={2025}
}

@inproceedings{reizenstein2021common,
  title={Common objects in 3d: Large-scale learning and evaluation of real-life 3d category reconstruction},
  author={Reizenstein, Jeremy and Shapovalov, Roman and Henzler, Philipp and Sbordone, Luca and Labatut, Patrick and Novotny, David},
  booktitle={Proceedings of the IEEE/CVF international conference on computer vision},
  pages={10901--10911},
  year={2021}
}

@inproceedings{zhu2023learning,
  title={Learning gabor texture features for fine-grained recognition},
  author={Zhu, Lanyun and Chen, Tianrun and Yin, Jianxiong and See, Simon and Liu, Jun},
  booktitle={Proceedings of the IEEE/CVF international conference on computer vision},
  pages={1621--1631},
  year={2023}
}

@inproceedings{li2018megadepth,
  title={Megadepth: Learning single-view depth prediction from internet photos},
  author={Li, Zhengqi and Snavely, Noah},
  booktitle={Proceedings of the IEEE conference on computer vision and pattern recognition},
  pages={2041--2050},
  year={2018}
}

\clearpage

\end{document}